\documentclass[3p,twocolumn]{elsarticle}
\usepackage{hyperref}
\journal{Journal of \LaTeX\ Templates}
\usepackage{amssymb}
\usepackage{graphicx}
\usepackage{textcomp}
\usepackage{amsmath,amssymb,amsfonts}
\usepackage{algorithmic}
\usepackage{graphicx}
\usepackage{caption}
\usepackage{textcomp}
\usepackage{color}
\usepackage[dvipsnames]{xcolor}
\usepackage{subfigure}
\usepackage{multirow}
\usepackage{rotating}
\usepackage{diagbox}
\usepackage{stfloats}
\usepackage{float}
\usepackage{bm}
\usepackage{booktabs}
\usepackage{float}
\usepackage{multirow}
\usepackage{soul}
\soulregister\cite7 
\soulregister\citep7 
\soulregister\citet7 
\soulregister\ref7 
\soulregister\pageref7 
\usepackage{appendix}
\usepackage{gensymb}

\begin{document}

\begin{frontmatter}

\title{Is the aspect ratio of cells important in deep learning? A robust comparison of deep learning methods for multi-scale cytopathology cell image classification: from convolutional neural networks to visual transformers.}

\author[a]{Wanli Liu}

\author[a]{Chen Li\corref{mycorrespondingauthor}}
\cortext[mycorrespondingauthor]{Corresponding author}
\ead{lichen201096@hotmail.com}

\author[a]{Md Mamunur Rahaman}

\author[b]{Tao Jiang}

\author[c]{Hongzan Sun}

\author[d]{Xiangchen Wu}

\author[a]{Weiming Hu}

\author[a]{Haoyuan Chen}

\author[a,e]{Changhao Sun}

\author[f]{Yudong Yao}

\author[g]{Marcin Grzegorzek}

\address[a]{Microscopic Image and Medical Image Analysis Group, College of Medicine and Biological Information Engineering, Northeastern University, Shenyang, 110169, China}

\address[b]{School of Control Engineering, Chengdu University of Information Technology, Chengdu, 610225, China}

\address[c]{Shengjing Hospital, China Medical University, Shenyang, 110001, China}

\address[d]{Suzhou Ruiguan Technology Company Ltd., Suzhou, 215000, China}

\address[e]{Shenyang Institute of Automation, Chinese Academy of Sciences, Shenyang, 110169, China}

\address[f]{Department of Electrical and Computer Engineering, Stevens Institute of Technology, Hoboken, NJ 07030, US}

\address[g]{Institute of Medical Informatics, University of Luebeck, Luebeck, Germany}

\begin{abstract}
Cervical cancer is a very common and fatal type of cancer in women. Cytopathology images are often used to screen for this cancer. Given that there is a possibility that many errors can occur during manual screening, a computer-aided diagnosis system based on deep learning has been developed. Deep learning methods require a fixed dimension of input images, but the dimensions of clinical medical images are inconsistent. The aspect ratios of the images suffer while resizing them directly. Clinically, the aspect ratios of cells inside cytopathological images provide important information for doctors to diagnose cancer. Therefore, it is difficult to resize directly. However, many existing studies have resized the images directly and have obtained highly robust classification results. To determine a reasonable interpretation, we have conducted a series of comparative experiments. First, the raw data of the SIPaKMeD dataset are pre-processed to obtain standard and scaled datasets. Then, the datasets are resized to 224 $\times$ 224 pixels. Finally, 22 deep learning models are used to classify the standard and scaled datasets. The results of the study indicate that deep learning models are robust to changes in the aspect ratio of cells in cervical cytopathological images. This conclusion is also validated via the Herlev dataset.
\end{abstract}


\end{frontmatter}

\section{Introduction\label{sec:Introduction}}
Cancer is an extremely fatal disease. Approximately 9.6 million people die from cancer every year~\cite{Bugdayci2019Roles}. According to the latest global cancer statistics, cervical cancer is the seventh most common cancer globally and ranks fourth in the incidence of cancer in women. In developing and low-income countries, the mortality rate due to cervical cancer is very high due to the lack of hygiene knowledge and access to medical facilities~\cite{ferlay2018global}. Human papillomavirus (HPV) is the leading cause of cervical cancer~\cite{Chaturvedi2012Epidemiology}. 

HPV is the most common sexually transmitted virus worldwide~\cite{Bernard2010Classification}. To date, approximately 200 types of HPV viruses have been discovered. Among these types, 13 are considered as high-risk types. Seventy percent of all cervical cancers are caused by HPV types 16 and 18~\cite{Crosbie2013Human,Cheng2005Human}. HPV type 16 is detected in 24\% of women, and 9\% of women are infected with HPV type 18~\cite{Crosbie2013Human,De2007Worldwide,Bosch2008Epidemiology}. The most common causes of cervical cancer are premature sexual intercourse, early pregnancy, intercourse with multiple partners, weak immune system, smoking, oral contraceptives, and improper menstrual hygiene. Common symptoms related to cervical cancer are abnormal vaginal bleeding, vaginal discharge, and moderate pain during sexual intercourse~\cite{vsarenac2019cervical}. Cervical cancer can be prevented if it is detected early and treated appropriately~\cite{saslow2012american}.

Cervical cytopathology is usually used to diagnose cervical malignancies. Given that the cost of the cervical cytopathological examination is relatively low, it is convenient for general investigation~\cite{saslow2012american,Davey2006Effect}. The traditional detection process involves collecting cells from the patient's cervical squamous column with a brush or spatula and then smearing the cells on a glass slide. A professional pathologist observes it under a microscope and determines the occurrence of any malignancies. However, manual screenings are difficult, time-consuming, and prone to artificial errors because thousands of cells are in different orientations and superimposed on each slide~\cite{GENCTAV2012Unsupervised}. Hence, the problem urgently requires a solution.

To solve the problems associated with manual screening, an automatic computer-aided diagnosis system (CAD) has been proposed to analyse the pap slices rapidly and accurately. Additionally, the computerised system can also process a large number of images, which is convenient for clinical follow-up, comparative research, and personalised medicine~\cite{Rahaman2020A}. Currently, deep learning techniques are widely used in computer vision, medical imaging, natural language processing, and other fields~\cite{lecun2015deep,Xing2016Robust}. In traditional CAD systems, the main limitation is that the features are manually extracted. This does not guarantee optimal classification results. However, the deep learning technique is an end-to-end method that can automatically learn features~\cite{goodfellow2016deep}. The performance of deep learning technology in image analysis is better than that of traditional machine learning-based methods. However, training a deep learning model requires a considerable number of labelled datasets~\cite{landau2019artificial}. A key question on deep learning is that although it provides excellent results, its characteristics are incomprehensible to humans~\cite{hinton2018deep}.

The deep learning programs, which are currently used in the field of medical image analysis, have proven to be the most efficient algorithms. A convolutional neural network (CNN) is a trendy deep architecture. It is widely used in image segmentation and classification~\cite{lecun2015deep}. A CNN model classifies normal and abnormal cells in the cell classification task by extracting hierarchical features from the original image~\cite{Lecun2010Convolutional}. Recently, a visual transformer (VT) that applies the transformer architecture for processing natural language processing (NLP) tasks to the field of computer vision was used. A VT model termed as vision transformer (ViT) exhibits better results than the latest convolutional network under the premise of pre-training with a large amount of data, and the computing resources required for training are significantly reduced. Nevertheless, when the amount of data is small, the effect may not be ideal~\cite{Dosovitskiy2020An}.

\begin{figure}[h]
	\centering
	  \includegraphics[scale=.37]{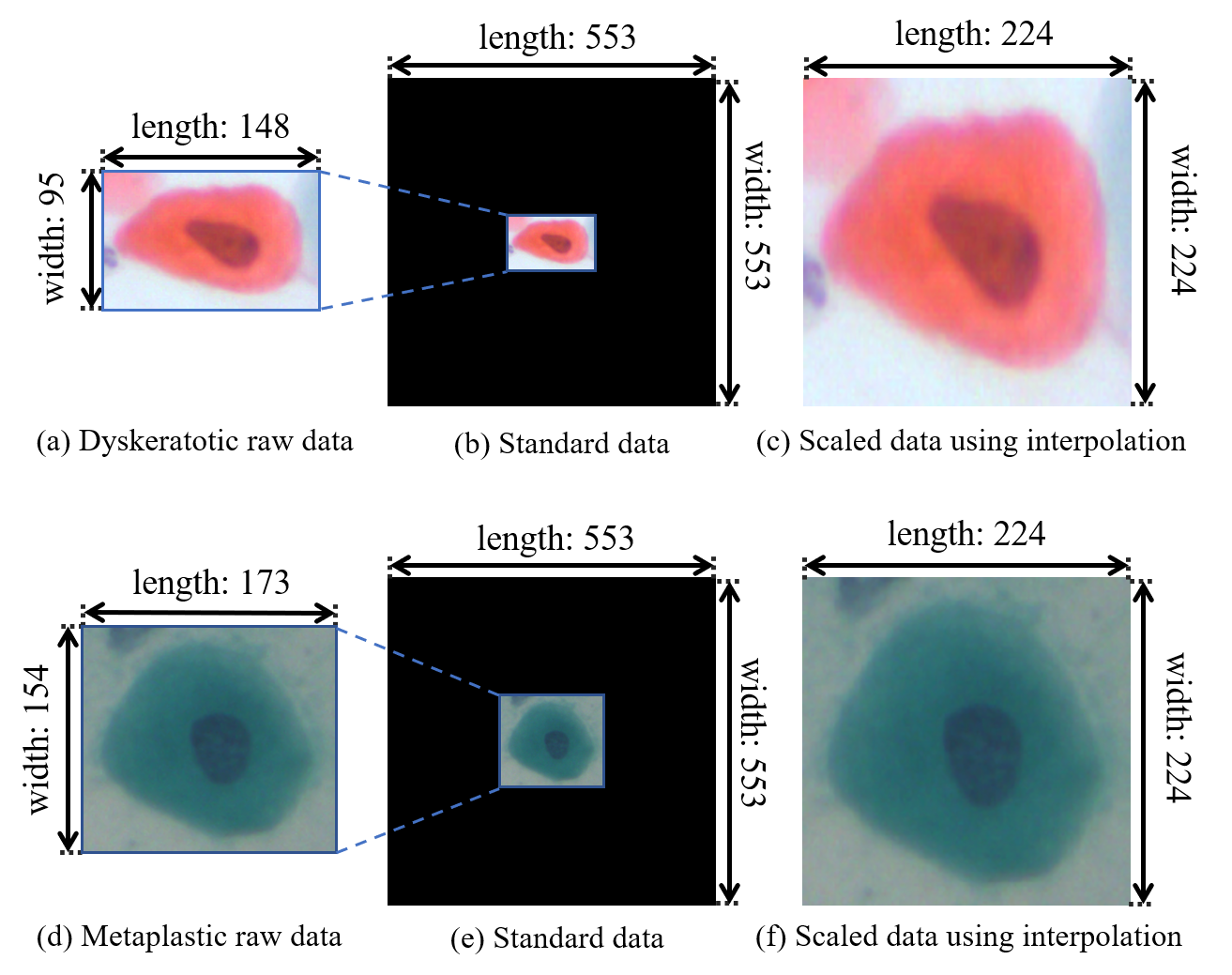}
	\caption{\centering{Visual comparisons of raw data, standard data, and scaled data.}}
	\label{FIG:resize}
\end{figure}

\begin{figure*}[h]
	\centering
	  \includegraphics[scale=.6]{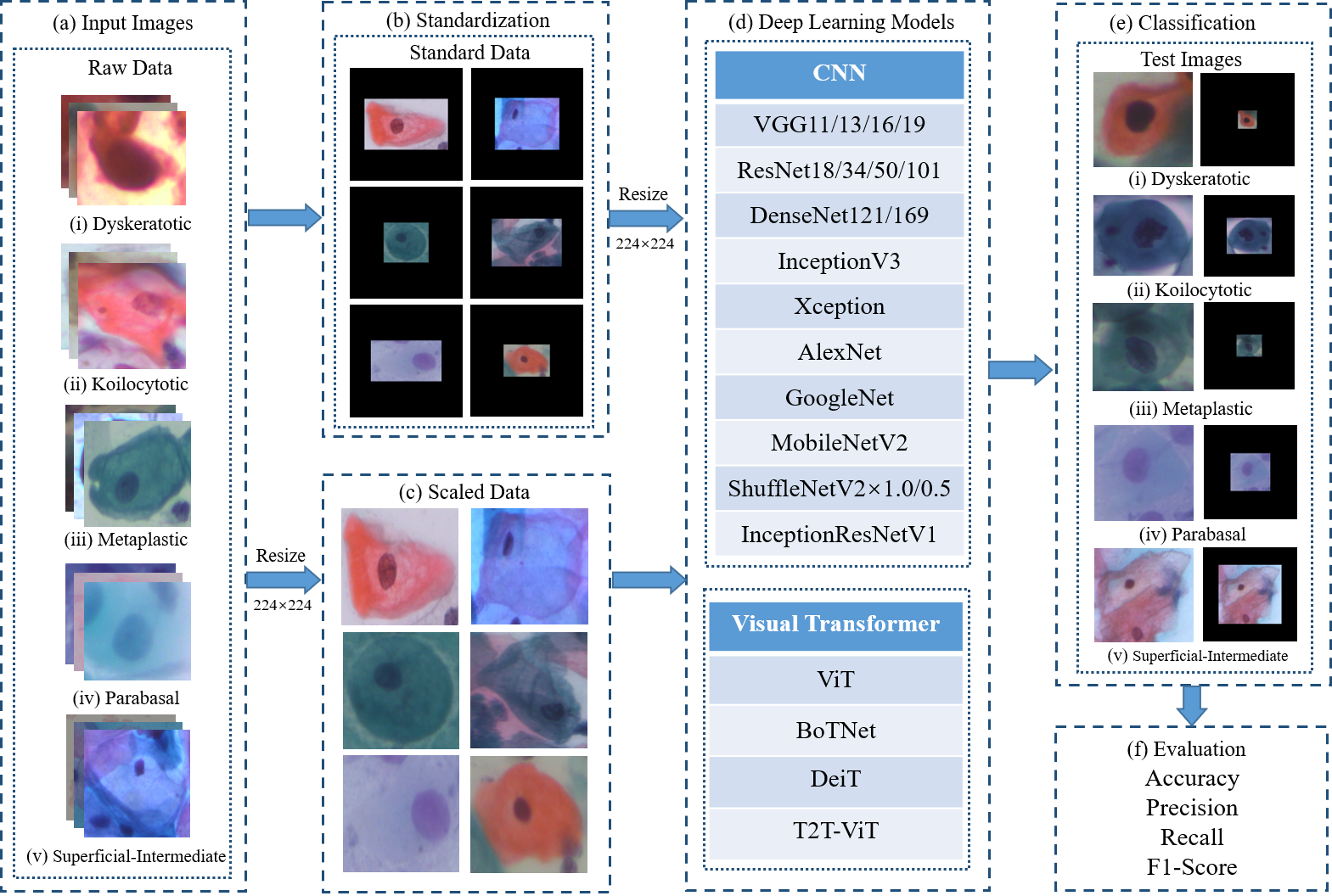}
	\caption{\centering{Workflow for comparing robustness of deep learning methods for multiscale cell image classification.}}
	\label{FIG:workflow}
\end{figure*}

The input of the deep learning models is consistent, but the dimensions of clinical cervical cell images are mostly inconsistent. The aspect ratio information are lost after resizing the images directly. A comparison of the raw data, standard data, and scaled data using interpolation is shown in Fig.~\ref{FIG:resize}. The scaled data are obtained after resizing the raw data to 224 $\times$ 224 pixels. It can be observed that the image aspect ratio ranges from 148 : 95 to 224 : 224 as a result of scaling up the dyskeratotic raw data. Hence, the change is significant. Moreover, the cell aspect ratio, which is the ratio between the cytoplasm and nucleus is also lost. However, the cell aspect ratio in cytopathological images provides important information for doctors to clinically diagnose cancer~\cite{zhao2017cervical}. According to the standard data in the figure, it can be observed that in the process of standardising the dyskeratotic raw data, this important information is not lost. Moreover, the resulting scaled data in the dyskeratotic class and metaplastic class are very similar after direct resizing, which contradicts the clinical explanations. Therefore, it is difficult to resize directly. However, most existing studies directly resize cell images, and their results are still robust. To find a reasonable explanation, in this study, we propose a robust comparison of deep learning methods for multiscale cell image classification tasks.

In this study, an experimental platform for the robustness comparison of deep learning methods in multiscale cell image classification tasks is developed. A workflow diagram of the proposed method is presented in Fig.~\ref{FIG:workflow}. First, raw data from the SIPaKMeD public dataset, consisting of 4049 cell images, are used in this study~\cite{plissiti2018sipakmed}. Second, the raw data are standardised and resized directly to obtain the standard and scaled data, respectively. Third, the scaled and standard data are resized to 224 $\times$ 224 pixels and used as training samples. Finally, the unseen test images are fed into the models for classification. The performance is evaluated by calculating the accuracy, precision, recall, and F1-Score on the validation and test sets.

In addition, 22 well-known deep learning models are selected for this experiment. Eighteen of the models are CNNs, including VGG11~\cite{simonyan2014very}, VGG13~\cite{simonyan2014very}, VGG16~\cite{simonyan2014very}, VGG19~\cite{simonyan2014very}, ResNet18~\cite{he2016deep}, ResNet34~\cite{he2016deep}, ResNet50~\cite{he2016deep}, ResNet101~\cite{he2016deep}, DenseNet121~\cite{huang2017densely}, Dense\-Net169~\cite{huang2017densely}, InceptionV3~\cite{Szegedy2016Rethinking}, Xception~\cite{Chollet2017Xception}, AlexNet~\cite{russakovsky2015imagenet}, GoogLeNet~\cite{Szegedy2015googlenet}, MobileNetV2~\cite{sandler2018mobilenetv2}, ShuffleNetV2$\times$1.0~\cite{ma2018shufflenet}, ShuffleNetV2$\times$0.5~\cite{ma2018shufflenet}, and InceptionResNetV1~\cite{szegedy2017inception}. The remaining four models are VTs, including ViT~\cite{Dosovitskiy2020An}, BoTNet~\cite{Srinivas2021Bottleneck}, DeiT~\cite{touvron2020training}, and T2T-ViT~\cite{yuan2021tokens}.

The structure of this paper is as follows: In Section~\ref{sec:Related Work}, we elaborate on related studies on deep learning for the classification of cervical cells. In Section~\ref{sec:Materials and Methods}, we describe the datasets, data pre-processing methods, and deep learning models. In Section~\ref{sec:Experiments and Analysis}, we explain the results of the comparative experiment and analysis of the experimental results, including the experimental settings and evaluation indicators.  In Section~\ref{sec:Conclusion and Future Work},  we summarise this study and propose future research directions.


\section{Related studies\label{sec:Related Work}}

In this section, related studies on deep learning in cervical cell image classification tasks are briefly introduced. For more detailed information, please refer to our review paper~\cite{Rahaman2020A}.


In~\cite{zhang2017deeppap}, deep learning and transfer learning were applied to the binary classification task of cervical cell images for the first time. In this study, the Herlev and HEMLBC datasets were used to evaluate the performance. All images were resized to 256 $\times$ 256 pixels. The HEMLBC dataset is a private database containing 2370 cervical cell images. The accuracy of the HEMLBC dataset was 98.6\%. The accuracy of the Herlev data set was 98.3\%.

In~\cite{hyeon2017automating}, a model that combines CNN and machine learning to classify cervical cell images was proposed. With respect to the images, the VGG network was used to extract features, and a support vector machine (SVM) was used for classification. In the aforementioned study, a private dataset containing 71344 cervical cell images was used. Furthermore, the original cell image was divided into several images of 224 $\times$ 224 pixels. The F1-Score of the classification result was 78\%.

In~\cite{taha2017classification}, the Herlev dataset was used wherein 70\% of the images were used for training and 30\% of the images were used for testing. Furthermore, images smaller than 227 $\times$ 227 pixels in the dataset were supplemented by blank areas. In the aforementioned study, the pre-trained AlexNet network was used for feature extraction. Then, the features were transferred to SVM for two classifications. The accuracy of this method was 99.19\%.

In~\cite{wieslander2017deep}, VGG16, and ResNet architecture networks were used to classify the CerviSCAN and Herlev datasets. CerviSCAN is a private database containing 12043 cervical cell images. In this study, the input dimension of the image was resized to 100 $\times$ 100 pixels. The F1-Score on VGGNet was 82\%, and F1-Score on ResNet was 83\%.

A cervical cell analysis system for detection, segmentation, and classification was proposed in~\cite{Gautam2018Considerations}. In this study, AlexNet was used for transfer learning to classify the Herlev dataset. They demonstrated that segmentation is not necessarily related to classification. The accuracy of the 2-class classification task and 7-class classification task was as high as 99.3\% and 93.75\%, respectively.

A comparative experiment using deep learning to classify cervical cell images was proposed in~\cite{promworn2019comparisons}. This study compared the performance of ResNet101, DenseNet161, Al\-exNet, VGG19\_bn, and SqueezeNet1\_1 on the Herlev dataset. Among the datasets, DenseNet161 exhibited the best performance with accuracy rates of 94.38\% and 68.54\% for 2-class classification and 7-class classification, respectively.

In~\cite{Nguyen2019Biomedical}, InceptionV3, ResNet152, and InceptionResNetV2 were used for feature concatenation and ensemble learning to classify the cervical cells. In this study, 2D Hela dataset, Pap smear, and Hep-2 cell image datasets were used for performance evaluation. All the images were resized to 256 $\times$ 256 pixels. The comprehensive result of multiple models in this method is better than the result of a single model, and the accuracy rate on the Herlev dataset was as high as 93.04\%.

In~\cite{Kurnianingsih2019Segmentation}, the pre-segmented cervical cell images were transferred into a compact version of the VGG for classification. This compact version has only seven layers, which in turn reduces the computational cost. In this study, the Herlev dataset was used for performance evaluation. The width of the image was resized to 200 pixels and was proportional to the length. The accuracy of the 2-class classification was 98.1\%. The accuracy of the 7-class classification was 95.9\%.

In~\cite{gv2019automatic}, a method for directly classifying whole slide image (WSI) cervical cell clusters without segmentation was proposed. In this study, a feature interpretation method was proposed based on principal component analysis (PCA) to visualise and understand the model. The method was evaluated on the SIPaKMeD and Herlev datasets, and WSI patches and single-cell images were resized to 224  $\times$ 224 and 80  $\times$ 80 pixels, respectively. The accuracy of this method was 96.37\% for WSI patches and 99.63\% for single-cell images.

In~\cite{xue2020application}, an ensemble transfer learning (ETL) framework for classifying cervical histopathological images was proposed. In this method, the transfer learning structures of InceptionV3, Xception, VGG16, and ResNet50 were developed. Then, an evolutionary learning strategy based on weighted voting was introduced. This method was verified on a dataset composed of 307 images, and the image was resized to 299 $\times$ 299 pixels and 224 $\times$ 224 pixels and input to the network. The highest accuracy rate obtained was 98.61\%.

In~\cite{dong2020inception}, a cell classification algorithm combined with InceptionV3 and artificial features was proposed. This method inherits the powerful learning ability of transfer learning to solve the problem of underfitting. In this method, the Herlev dataset was used to classify cervical cell images. The images of the dataset were filled with zero to maintain an aspect ratio of 1:1, and then they were resized to 299 $\times$ 299 pixels. The accuracy rate of this study exceeded 98\%.

In~\cite{win2020computer}, a system for cervical cancer detection and classification was proposed. This method used an ensemble classifier that combines linear discriminant, SVM, \emph{k-nearest} neighbour, boosted trees, and bagged trees for classification. In the SIPaKMeD dataset, the accuracy rate of the 2-class classification was 98.27\%, and the accuracy rate of the 5-class classification was 94.09\%.

In~\cite{khamparia2020internet}, a novel internet of health things (IoHT)-driven deep learning framework for the detection and classification of cervical cancer images was proposed. This study uses pre-trained CNN models for feature extraction and machine learning models for classification. The Herlev dataset was used to evaluate the performance. With the participation of the random forest classifier, the CNN pre-training model ResNet50 realized a high accuracy rate of 97.89\%.

In~\cite{khamparia2020dcavn}, a method combining a convolutional network and variable autoencoder for cervical cell classification was proposed. The use of a variable encoder reduced the dimensionality of the data for further processing with the participation of the softmax layer. This method uses the Herlev dataset for evaluation, and the images were resized at different scales to prevent further distortions. This study realized a maximum accuracy of 99.4\%.

A cervical cancer cell detection and classification system based on CNNs was proposed in~\cite{ghoneim2020cervical}. In this method, the CNN model was used to extract features and then used as a classifier based on an extreme learning machine for classification. The CNN model input corresponded to an image of 224 $\times$ 224 pixels. The Herlev dataset was used to evaluate the performance, and the accuracy rates for the 2-class and 7-class classification tasks were 99.5\% and 91.2\%, respectively.

In~\cite{mamunur2021deepcervix}, a hybrid deep feature fusion method based on deep learning was proposed to accurately classify cervical cells. In this study, the SIPaKMeD dataset was used to evaluate the performance of the model. All the images were resized to 224 $\times$ 224 pixels for 2-class, 3-class, and 5-class classification tasks, and the accuracy rates were 99.85\%, 99.38\%, and 99.14\%, respectively.

In~\cite{shi2021cervical}, a classification method for cervical cells based on a graph convolutional network (GCN) was proposed. In this study, the CNN features of all cervical cell images were clustered to construct a graph structure. Then, the GCN was applied to generate relationship-aware features. The GCN features were merged to enhance the distinguishing ability of the CNN features. This study demonstrated the feasibility and effectiveness of the SIPaKMeD dataset. Each image of this dataset was resized to 80 $\times$ 80 pixels with an accuracy rate of 98.37\%.

In~\cite{manna2021fuzzy}, an ensemble method using InceptionV3, Xception, and DenseNet169, three pre-trained CNN models for cervical cell image classification, was proposed. In this method, a classifier fusion was used based on fuzzy levels. The SIPaKMeD and Mendeley datasets were used for the evaluation. The accuracy rates of the 2-class and 5-class classification tasks on the SIPaKMeD dataset were 98.55\% and 95.43\%, respectively. The accuracy rate realized using the Mendeley dataset was 99.23\%.

In~\cite{bhatt2021cervical}, a new method using transfer learning and progressive resizing technology was proposed. The SIPaKMeD dataset was used for the evaluation. In this method, the model was iteratively trained by gradually increasing the image dimension from 224 $\times$ 224 pixels to 256 $\times$ 256, 512 $\times$ 512, and 1024 $\times$ 1024 pixels. The accuracy rate of the WSI image multi-classification was as high as 99.70\%.

In~\cite{basak2021cervical}, a fully automated framework for cytological image classification was proposed. The framework extracts deep features from CNN models and uses evolutionary optimisation algorithms and grey wolf optimisers to select the optimal feature subset. Finally, the SVM classifier was used for classification. Specifically, Mendeley, Herlev, and SIPaKMeD datasets were used for the evaluation, and the classification accuracy rates were 99.47\%, 98.32\%, and 97.87\%, respectively.

\section{Materials and methods\label{sec:Materials and Methods}}

\subsection{Datasets}

\subsubsection{SIPaKMeD}

\begin{figure*}[h]
	\centering
	  \includegraphics[scale=0.58]{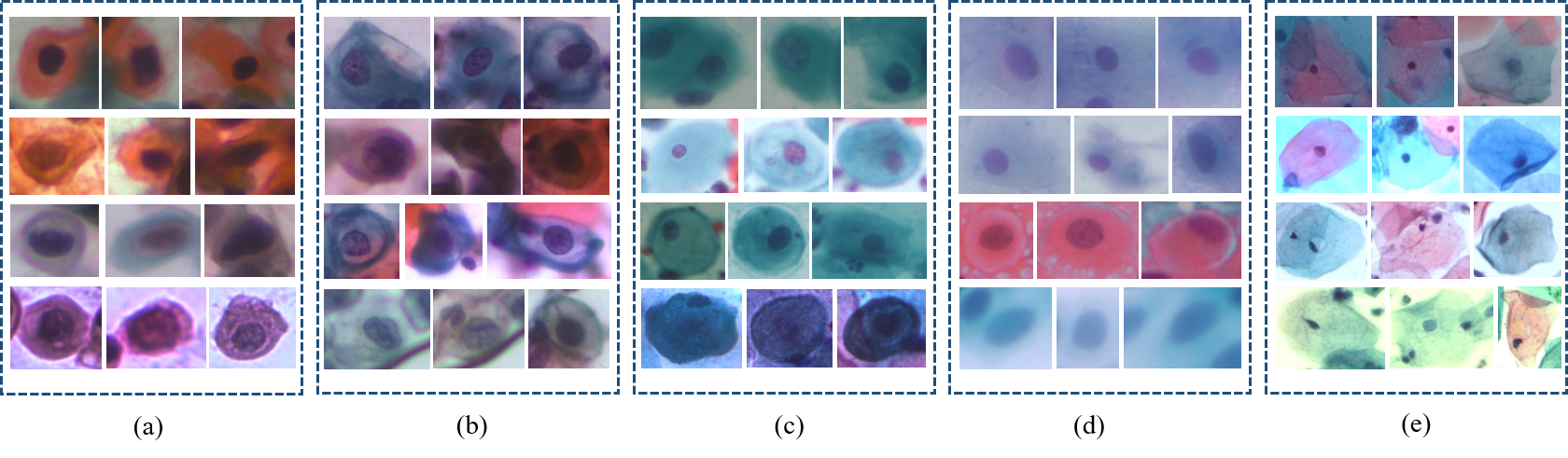}
	\caption{Example of SIPaKMeD dataset: (a) Dyskeratotic, (b) Koilocytotic, (c) Metaplastic, (d) Parabasal, and (e) Superficial-Intermediate.}
	\label{FIG:SIPaKMeD}
\end{figure*}

A publicly available SIPaKMeD dataset is used to compare the performance of the deep learning models. The database contains 4049 separated cervical cell images. All the images are cropped from clustered cell images obtained via a CCD camera. These images are classified into five categories: dyskeratotic, koilocytotic, metaplastic, parabasal, and superficial intermediate. Some examples of the SIPaKMeD dataset are shown in Fig.~\ref{FIG:SIPaKMeD}~\cite{plissiti2018sipakmed}. In this figure, It is observed that the colour of the cells in the same category is different. The colour information in the cytopathology images is not significant, but it is very important for the histopathology images.

In each category of the dataset, 60\% of the data are randomly selected for training, 20\% for validation, and the remaining data are used for testing. Then, deep learning models are used to perform 5-class classification. The training, validation, and test sets are listed in Table~\ref{SIPaKMeD_data_arrangement}.

To further validate the model performance, we perform five-fold cross-validation, as described in Extended Experiment subsection~\ref{sec:SIPaKMeD_cross_validation}.

In the Extended Experiment subsection~\ref{sec:SIPaKMeD_aug_extend}, the aforementioned training and validation sets of the SIPaKMeD dataset are augmented, and the data increases four times by rotating 180\degree, flipping left and right, and flipping up and down. Then, the augmented data are standardised and resized directly to obtain the standard and scaled data, respectively. Augmented scaled data and standard data are used to train the models. The arrangement of the data is presented in Table~\ref{SIPaKMeD_data_arrangement}.

\begin{table}[htb]
\caption{Arrangement of SIPaKMeD data and augmented data. (D: Dyskeratotic, K: Koilocytotic, M: Metaplastic, P: Parabasal, S: Superficial-Intermediate. The upper part of the table is the SIPaKMeD data arrangement, and the lower part is the SIPaKMeD augmented data arrangement.)}\label{SIPaKMeD_data_arrangement}
\scalebox{0.9}{
\begin{tabular}{@{} ccccccc@{} }
\toprule
Data/Class & D & K & M & P & S & Total \\
\midrule
Train & 488 & 495 & 476 & 473 & 499 & 2431\\
Val & 163 & 165 & 159 & 157 & 166 & 810\\
Test & 162 & 165 & 158 & 157 & 166 & 808\\
Total & 813 & 825 & 793 & 787 & 831 & 4049\\
\midrule
Train & 1952 & 1980 & 1904 & 1892 & 1996 & 9724\\
Val & 652 & 660 & 636 & 628 & 664 & 3240 \\
Test & 162 & 165 & 158 & 157 & 166 & 808\\
Total & 2766 & 2805 & 2698 & 2677 & 2826 & 13772\\
\bottomrule
\end{tabular}}
\end{table}

\subsubsection{Herlev}

\begin{figure*}
\centering
\includegraphics[scale=0.51]{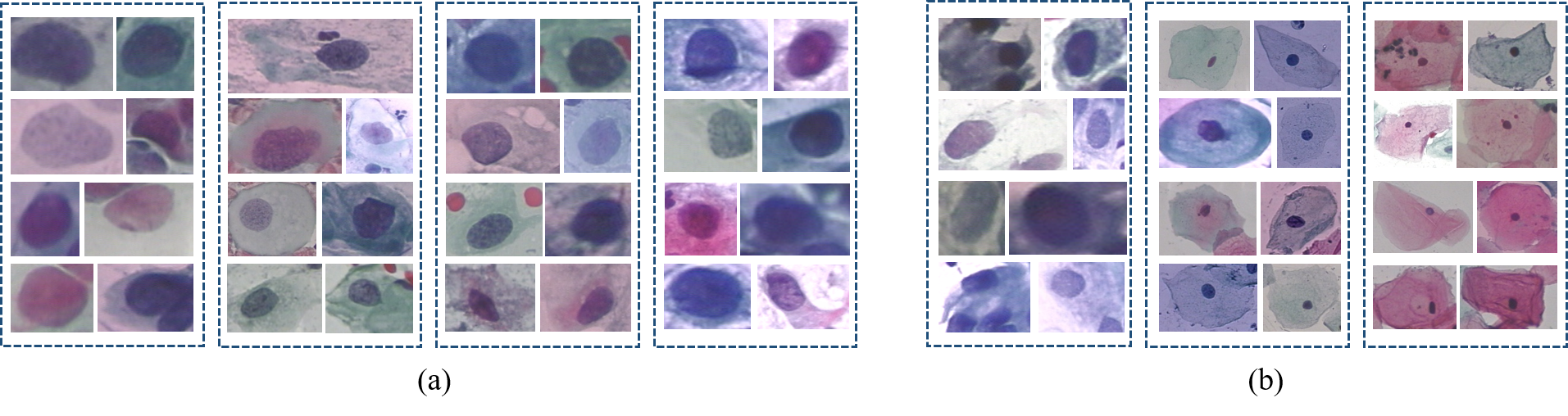}
\caption{Example of Herlev dataset: (a) \textbf{Abnormal} (From left to right: Carcinoma \emph{in situ}, Mild dysplasia, Moderate dysplasia, Severe dysplasia.), (b) \textbf{Normal} (From left to right: Columnar, Intermediate squamous, Normal squamous.).}
\label{FIG:Herlev}
\end{figure*}

The Herlev dataset contains 917 cell images, including seven categories: normal squamous, intermediate squamous, columnar, mild dysplasia, moderate dysplasia, severe dysplasia, and carcinoma \emph{in situ}. Furthermore, these seven categories can also be further classified into normal and abnormal types, where the normal type includes normal squamous, intermediate squamous, and columnar with 242 images, and the abnormal type includes carcinoma \emph{in situ}, mild dysplasia, moderate dysplasia, severe dysplasia with 675 images, as shown in Fig.~\ref{FIG:Herlev}~\cite{jantzen2005pap}.

In Extended Experiment subsection~\ref{Herlev_extend}, the Herlev dataset is used for the two-class classification tasks. In each category of the dataset, 60\% of the data are randomly selected for training, 20\% for validation, and 20\% for testing. Given that the amount of data is relatively small, the data is increased by four times by rotating 180\degree, flipping left and right, and flipping up and down. Then, the augmented data are standardised and resized directly to obtain the standard and scaled data, respectively. Augmented scaled data and standard data are used as the input of the models for classification. The arrangement of augmented data is presented in Table~\ref{Herlev_augmented_data_arrangement}.

\begin{table}[htbp]
\caption{Arrangement of Herlev augmented data.}\label{Herlev_augmented_data_arrangement}
\centering
\scalebox{1}{
\begin{tabular}{@{} cccc@{} }
\toprule
Data/Class & Normal & Abnormal  & Total \\
\midrule
Train        & 584 & 1620 & 2204 \\
Val           & 192 & 540 & 732  \\
Test          & 48 & 135 & 183 \\
Total         & 824 & 2295 & 3119 \\
\bottomrule
\end{tabular}}
\end{table}

\subsection{Data pre-processing}

\begin{figure}[htbp]
	\centering
	  \includegraphics[scale=0.7]{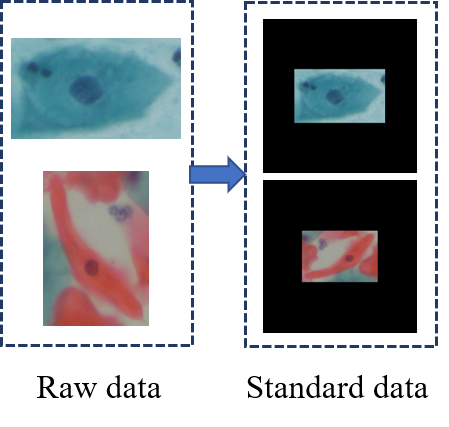}
	\caption{Data pre-processing examples.}
	\label{FIG:standard_data}
\end{figure}

The dimensions of cervical cell images in the dataset are inconsistent, but the inputs required by the deep learning models are consistent. MATLAB is used to standardise and centralise the cervical cell images (raw data) in the dataset to determine whether the aspect ratios of cells in the image impact the performance of the models. The standard operation is to check whether the width of the raw data exceeds the length; If the width exceeds the length, then the raw data is rotated 90\degree counterclockwise, else no processing is performed. The centralisation operation is performed to find the longest side length in all the images, and then zero is used to supplement all images to square matrices of that length. The standardised and centralised images are termed as the ``standard data" and some examples are shown in Fig.~\ref{FIG:standard_data}.

\subsection{Deep learning models}

The robustness comparison experiment uses deep learning models to classify cervical cell images. First, the training and validation sets generated by the scaled data and standard data are used to train the models. The test set is then used to evaluate the performance of the models. The two classification results are compared and analysed using the obtained evaluation indicators to determine whether the aspect ratio of cells affects the model performance. Twenty-two deep learning models are used in this experiment. Eighteen of the models correspond to CNNs, including VGG11~\cite{simonyan2014very}, VGG13~\cite{simonyan2014very}, VGG16~\cite{simonyan2014very}, VGG19~\cite{simonyan2014very}, ResNet18~\cite{he2016deep}, ResNet34~\cite{he2016deep}, ResNet50~\cite{he2016deep}, ResNet101~\cite{he2016deep}, DenseNet121~\cite{huang2017densely}, DenseNet169~\cite{huang2017densely}, InceptionV3~\cite{Szegedy2016Rethinking}, Xc\-eption~\cite{Chollet2017Xception}, AlexNet~\cite{russakovsky2015imagenet}, GoogLeNet~\cite{Szegedy2015googlenet}, MobileNetV2~\cite{sandler2018mobilenetv2}, ShuffleNetV2$\times$1.0~\cite{ma2018shufflenet}, ShuffleNetV2$\times$0.5~\cite{ma2018shufflenet}, and InceptionResNetV1~\cite{szegedy2017inception}. Four of the models correspond to VTs, including ViT~\cite{Dosovitskiy2020An}, BoTNet~\cite{Srinivas2021Bottleneck}, DeiT~\cite{touvron2020training}, and T2T-ViT~\cite{yuan2021tokens}.

\begin{figure*}[h]
	\centering
	  \includegraphics[scale=0.73]{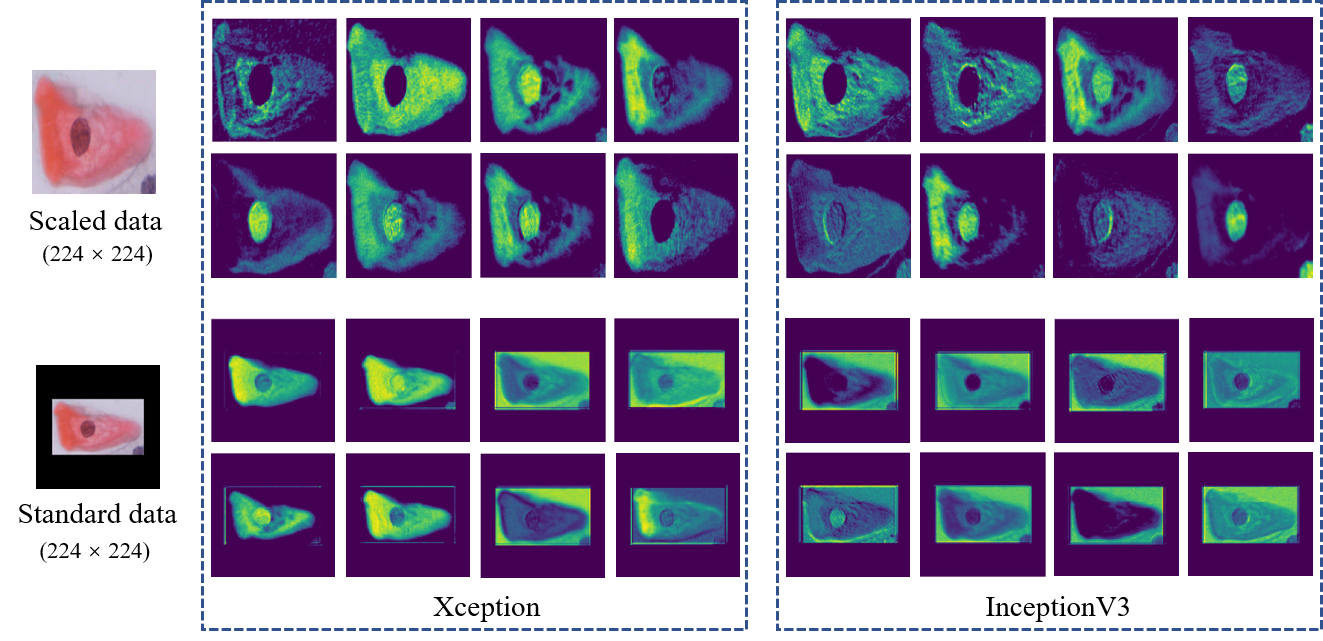}
	\caption{Example of feature maps.}
	\label{FIG:feature_map}
\end{figure*}

The references, parameters, top-1 accuracy on the ImageNet dataset, and major remarks of the 22 models are presented in the Appendix (Table~\ref{model_introduce}).

To determine how the CNN models extract features to classify images, we use Xception and InceptionV3 as examples. The input images correspond to scaled and standard data. Feature maps extracted from the second convolution block in the network are shown in Fig.~\ref{FIG:feature_map}. It is determined that the amount of visual information in the nucleus is more than that in the cytoplasm in this figure, and this provides a higher degree of discrimination.

\section{Experiments and analysis\label{sec:Experiments and Analysis}}

\subsection{Experimental setup}

This comparative experiment is conducted on a local computer. The running memory of the computer is 32 GB. The computer uses Win10 Professional operating system, and it is equipped with an 8 GB NVIDIA GeForce RTX 2080 GPU. Python 3.7, Pytorch 1.8.0, and Torchvision 0.9.0 are configured on this computer. For all the models, the image input dimension is 224 $\times$ 224 $\times$ 3. The learning rate is 0.0002. The epoch is set to 100, and the AdamW optimiser is employed. The batch size of the training set is 32.

\subsection{Evaluation method}

It is very important to select an appropriate evaluation method in a comparative experiment. Specifically, precision, recall, F1-Score, and accuracy are the most commonly used and standard evaluation methods in classification experiments~\cite{xie2015beyond,sukumar2016computer}. By considering the positive–negative binary classification as an example, true positives (TP) correspond to the number of positive samples that are accurately predicted. The number of negative samples predicted as positive samples is termed as false positives (FP). The number of positive samples predicted as negative samples is termed false negatives (FN). True negatives (TN) is the number of negative samples that are accurately predicted. Precision is the proportion of TP among all the positive predictions. Recall is the ratio of the predicted positive examples to the total number of actual positive examples. The F1-Score combines precision and recall. Accuracy is the ratio of the number of correct predictions to the total number of samples. The equations for the four evaluation methods are listed in Table~\ref{tbl3}.

\begin{table}[htbp!]
\caption{Evaluation metrics}\label{tbl3}
\renewcommand\arraystretch{1.5}
\centering
\scalebox{1.1}{
\begin{tabular}{@{} cc@{} }
\toprule
Assessment & Formula \\
\midrule
Precision ($P$) & $\rm \frac{TP}{TP + FP}$ \\
Recall ($R$) & $\rm \frac{TP}{TP + FN}$ \\
F1-Score& $ 2 \times \frac{P \times R}{P + R}$ \\
Accuracy&$\rm \frac{TP+TN}{TP + TN + FP + FN}$ \\
\bottomrule
\end{tabular}}
\end{table}

\subsection{Experimental results and analysis}

The average precision, average recall, average F1-Score, and accuracy are used to evaluate the models. The 5-class classification results are presented in Table~\ref{tbl5}.

The closed test sets of the scaled data and standard data are used to evaluate the generalisation ability of the models (Table~\ref{tbl5}). The CNNs that use the scaled test set, GoogLeNet, realises the highest test accuracy of 96.03\%. The test accuracy of VGG19 is the lowest among all CNNs with an accuracy of 88.11\%. In VTs using the scaled test set, DeiT realises the highest test accuracy of 95.42\%. BoTNet exhibits the lowest accuracy rate (84.77\%). For the CNNs, the highest classification accuracy of 95.17\% is obtained by DenseNet169 by using the standard data as a test set. The test accuracy of ShuffleNetV2$\times$0.5 is the lowest among the CNNs with an accuracy of 90.34\%. Among all VTs using the standard data as a test set, DeiT has the highest test accuracy at 94.18\%. The lowest accuracy rate of 88.36\% is obtained using T2T-ViT.

Table~\ref{tbl6} compares the accuracy of the test set using scaled data and standard data. The test accuracy of the standard data corresponds to the baseline. In this table, the accuracies of 12 models, including VGG11, ResNet18, ResNet34, DenseNet121, DenseNet169, Xception, GoogLeNet, MobileNetV2, ShuffleNetV2$\times$0.5, ViT, DeiT, and T2T-ViT, tested on scaled data are higher than that of the standard data. The test accuracies of ten models, including VGG13, VGG16, VGG19, ResNet50, ResNet101, InceptionV3, AlexNet, ShuffleNetV2$\times$1.0, InceptionResNetV1, and BoTNet are reduced. Furthermore, T2T-ViT exhibits the highest  increase of 3.34\% and a ratio of 3.78\%. BoTNet exhibits the highest decrease of 6.81\%, and a ratio of 7.44\%. After the calculation, the average change in accuracy is -0.31\%. The average absolute value of the change in accuracy is 1.76\%, and the standard deviation of the absolute value of the changes is 1.41\%. The average ratio of the changes in accuracy is -0.33\%. The average of the absolute value of the change ratios is 1.90\%, and the standard deviation is 1.55\%.

It is observed that the accuracies of certain models increase and decrease.	

As shown in Table~\ref{tbl6}, when the number of layers in the VGG network increases, the classification performance of the scaled data is lower than that of the standard data. When the number of layers of the ResNet network increases, the classification performance of the scaled data is lower than that of the standard data. In the DenseNet series, a direct connection is introduced between any two layers with the same feature map size. This can lead to better performance of the scaled data. InceptionV3's poor performance on scaled data is due to the fact that one large two-dimensional convolution is replaced with two small one-dimensional convolutions. As an improved version of InceptionV3, the performance of Xception on the scaled data is higher than that on the standard data. Furthermore, AlexNet, which introduces the ReLu activation function, performs worse in scaled data processing. GoogLeNet introduces the inception structure and adopts the Hebbian principle and multiscale processing, which performs better on scaled data processing. Lightweight networks, such as MobileNet and ShuffleNet, have more advantages in processing scaled data.

In VTs, models have a better effect on scaled data than standard data. This is because the transformer is more focused on global information and lacks local perception.

In the case of classifying the scaled data with missing aspect ratio information, the classification effects of some models are slightly lower than that of the standard data. However, the decrease in classification effects is not high. Even, the effects of some models have been improved. This shows that the deep learning method is very robust to changes in the aspect ratio of cells in cervical cytopathological images.

To further investigate whether the results are stable, the following extended experiments are conducted.

\begin{table*}
\caption{Classification performance of models on the \textbf{test set} on the SIPaKMeD dataset.}\label{tbl5}
\centering
\scalebox{0.72}{
\begin{tabular}{@{}ccccccccccc@{} }
\toprule
 &\multicolumn{4}{c}{Scaled data} & &  & \multicolumn{4}{c}{Standard data} \\
 \midrule
Models           & Avg. P (\%) & Avg. R (\%)& Avg. F1 (\%) & Accuracy (\%) &&& Avg. P (\%) & Avg. R (\%) &Avg. F1 (\%) &  Accuracy (\%)\\
 \midrule
VGG11                      & 93.70  & 93.60  & 93.60   & 93.56  &&& 93.30  & 92.80  & 92.90   & 92.82\\
VGG13                      & 91.60  & 91.20  & 91.30   & 91.21  &&& 93.50  & 93.20  & 93.30   & 93.19\\
VGG16                      & 89.80  & 89.20  & 89.30   & 89.23   &&& 92.80 & 92.90  & 92.80   & 92.82\\
VGG19                      & 88.60  & 88.10  & 88.30   & 88.11  &&& 91.00   & 90.90 & 90.90   & 90.84\\
ResNet18                  & 95.60  & 95.50  & 95.60  & 95.54    &&& 93.80  & 93.80  & 93.80   & 93.81\\
ResNet34                  & 94.30  & 94.00   & 94.10   & 94.05   &&& 94.10  & 94.00   & 94.00    & 93.93\\
ResNet50                  & 93.10  & 92.80  & 92.80   & 92.82    &&& 94.90 & 94.70  & 94.70   & 94.67\\
ResNet101                 & 92.30  & 92.00  & 92.00    & 91.95   &&& 93.70  & 93.70  & 93.60   & 93.68\\
DenseNet121               & 95.80  & 95.70  & 95.70   & 95.66   &&& 95.20  & 94.90  & 95.00    & 94.92\\
DenseNet169               & 95.80  & 95.70  & 95.70   & 95.66    &&& 95.30  & 95.20  & 95.20   & \textbf{95.17}\\
InceptionV3               & 93.20  & 93.00   & 93.10   & 92.94   &&& 94.00   & 93.70  & 93.70   & 93.68\\
Xception                   & 96.10  & 95.90  & 95.90   & 95.91  &&& 95.20  & 94.50  & 94.60   & 94.55\\
AlexNet                    & 92.00   & 91.90  & 91.90   & 91.83    &&& 93.50  & 93.50 & 93.40   & 93.44\\
GoogLeNet                  & 96.30 & 96.00   & 96.10   &  \textbf{96.03}  &&& 94.30  & 94.10  & 94.10   & 94.05\\
MobileNetV2               & 93.80  & 93.70  & 93.70   & 93.68   && &93.10  & 93.20  & 93.20   & 93.19\\
ShuffleNetV2$\times$1.0          & 92.30  & 91.90  & 91.90   & 91.83   &&& 92.80  & 92.80  & 92.70   & 92.69\\
ShuffleNetV2$\times$0.5          & 92.70  & 92.30  & 92.40   & 92.32    &&& 90.40  & 90.40  & 90.40 & 90.34\\
InceptionResNetV1         & 93.80  & 93.70  & 93.70   & 93.68   &&& 94.70  & 94.60  & 94.60   & 94.55\\
ViT                        & 93.60  & 93.60 & 93.60   & 93.56    &&& 92.10  & 91.90  & 91.90   & 91.83\\
BoTNet                     & 85.80  & 84.80  & 84.90   & 84.77     &&& 91.50  & 91.60  & 91.50   & 91.58\\
DeiT                     &  95.50 &  95.40 &  95.50  &  \textbf{95.42}   &&& 94.20  & 94.30  &  94.20  & \textbf{94.18}\\
T2T-ViT                     &  91.90 & 91.70  &  91.70 & 91.70   && & 88.30 & 88.40  &  88.30  & 88.36 \\
\bottomrule
\end{tabular}}
\end{table*}

\begin{table}
\caption{Comparison table of scaled data test results and standard data test results on SIPaKMeD dataset. (\textbf{Up/down:} The accuracy of the model trained on the scaled data is higher/lower than that of the standard data. \textbf{Change:} The change value of the accuracy of the model trained on the scaled data relative to that of the standard data. \textbf{Ratio:} The ratio of the change value to the accuracy of the standard data. \textbf{Average (abs)} and \textbf{Standard deviation (abs)}  are calculated from the absolute values of change and ratio of 22 models.)}\label{tbl6}
\centering
\scalebox{0.75}{
\begin{tabular}{@{} cccc@{} }
\toprule
 Models            & Up/down & Change (\%) & Ratio (\%) \\
\midrule
VGG11             & up   & $+$0.74  & $+$0.80  \\
VGG13             & down & $-$1.98 & $-$2.13 \\
VGG16             & down & $-$3.59 & $-$3.87 \\
VGG19             & down & $-$2.73 & $-$3.01 \\
ResNet18          & up   & $+$1.73  & $+$1.84  \\
ResNet34          & up   & $+$0.12  &$+$0.13  \\
ResNet50          & down & $-$1.85 & $-$1.95 \\
ResNet101        & down &$-$1.73 &$-$1.85\\
DenseNet121    & up   & $+$0.74  & $+$0.78   \\
DenseNet169    & up   & $+$0.49  & $+$0.52  \\
InceptionV3      & down & $-$0.74 & $-$0.79 \\
Xception           & up   & $+$1.36  & $+$1.44  \\
AlexNet             & down & $-$1.61 & $-$1.72 \\
GoogLeNet        & up   & $+$1.98  & $+$2.11  \\
MobileNetV2      & up   & $+$0.49  & $+$0.53 \\
ShuffleNetV2$\times$1.0  & down & $-$0.86 & $-$0.93 \\
ShuffleNetV2$\times$0.5   & up   & $+$1.98  & $+$2.19  \\
InceptionResNetV1  & down & $-$0.87 & $-$0.92  \\
ViT                & up   & $+$1.73  & $+$1.88  \\
BoTNet          & down & \textbf{$-$6.81} & \textbf{$-$7.44} \\
DeiT              & up & $+$1.24 & $+$1.31 \\
T2T-ViT          & up   & \textbf{$+$3.34} & \textbf{$+$3.78} \\
Average           &  down   & $-$0.31  & $-$0.33  \\
Average (abs)    &   -  & 1.76 & 1.90  \\ 	 		 	
Standard deviation (abs) & -  & 1.41   &  	1.55   \\ 
\bottomrule
\end{tabular}}
\end{table}

\subsection{Extended experiment}

\subsubsection{Extended experiment on SIPaKMeD dataset: Five-fold cross-validation\label{sec:SIPaKMeD_cross_validation}}

\begin{table*}
\caption{Classification performance of models of \textbf{five-fold cross-validation} experiment on the SIPaKMeD dataset.}\label{cross_val_val}
\centering
\scalebox{0.72}{
\begin{tabular}{@{} ccccccccccc@{} }
\toprule
 &\multicolumn{4}{c}{Scaled data} &  & & \multicolumn{4}{c}{Standard data} \\
\midrule
Models           & Avg. P (\%) & Avg. R (\%)& Avg. F1 (\%) & Accuracy (\%) &&& Avg. P (\%) & Avg. R (\%) &Avg. F1 (\%) &  Accuracy (\%)\\
\midrule
VGG11             & 84.58 & 82.88 & 82.42 & 82.79 &  &  & 86.04 & 84.74 & 84.40 & 84.69 \\
VGG13             & 85.56 & 84.84 & 84.76 & 84.79 &  &  & 87.40 & 86.42 & 86.10 & 86.40 \\
VGG16             & 82.20 & 79.26 & 78.52 & 79.27 &  &  & 84.18 & 82.44 & 81.78 & 82.40 \\
VGG19             & 82.10 & 80.38 & 80.18 & 80.37 &  &  & 85.90 & 85.04 & 85.14 & 84.99 \\
ResNet18          & 87.68 & 86.46 & 85.94 & 86.42 &  &  & 88.10 & 87.62 & 87.26 & 87.50 \\
ResNet34          & 87.16 & 86.28 & 85.72 & 86.23 &  &  & 87.76 & 86.86 & 86.62 & 86.79 \\
ResNet50          & 84.18 & 82.22 & 81.56 & 82.25 &  &  & 88.00 & 87.30 & 87.24 & 87.26 \\
ResNet101         & 83.74 & 82.04 & 81.60 & 81.98 &  &  & 86.56 & 86.50 & 86.30 & 86.39 \\
DenseNet121       & 85.06 & 83.60 & 82.78 & 83.58 &  &  & 87.50 & 86.86 & 86.62 & 86.81 \\
DenseNet169       & 86.76 & 85.72 & 85.36 & 85.65 &  &  & 88.28 & 86.70 & 86.06 & 86.67 \\
InceptionV3       & 85.44 & 83.66 & 83.46 & 83.65 &  &  & 87.12 & 86.18 & 85.98 & 86.15 \\
Xception          & 89.06 & 87.98 & 87.84 &  \textbf{87.95} &  &  & 89.76 & 88.94 & 88.76 &  \textbf{88.83} \\
AlexNet           & 83.20 & 81.14 & 80.84 & 81.09 &  &  & 86.26 & 85.16 & 84.76 & 85.11 \\
GoogLeNet         & 87.18 & 85.22 & 84.82 & 85.18 &  &  & 88.94 & 87.58 & 87.46 & 87.51 \\
MobileNetV2       & 86.76 & 84.96 & 84.76 & 84.94 &  &  & 86.50 & 85.38 & 85.18 & 85.33 \\
ShuffleNetV2$\times$1.0  & 85.92 & 84.54 & 83.82 & 84.52 &  &  & 85.92 & 84.56 & 84.00 & 84.49 \\
ShuffleNetV2$\times$0.5  & 83.54 & 82.58 & 82.26 & 82.52 &  &  & 84.58 & 83.16 & 82.66 & 83.08 \\
InceptionResNetV1 & 87.02 & 85.20 & 84.96 & 85.23 &  &  & 88.00 & 87.22 & 87.04 & 87.16 \\
ViT               & 85.72 & 83.36 & 82.70 & 83.33 &  &  & 82.18 & 81.30 & 80.10 & 81.26 \\
BoTNet            & 79.60 & 78.06 & 76.44 & 78.05 &  &  & 82.50 & 81.56 & 81.12 & 81.48 \\
DeiT              & 86.32 & 85.16 & 84.74 &  \textbf{85.16} &  &  & 86.68 & 85.14 & 84.66 & 85.09 \\
T2T-ViT           & 85.64 & 83.58 & 83.28 & 83.61 &  &  & 85.80 & 85.24 & 84.98 &  \textbf{85.16}\\

\bottomrule
\end{tabular}}
\end{table*}

\begin{table}
\caption{Comparison table of scaled data validation results and standard data validation results of five-fold cross-validation experiment on the SIPaKMeD dataset. (\textbf{Up/down:} The accuracy of the model trained on the scaled data is higher/lower than that of the standard data. \textbf{Change:} The change value of the accuracy of the model trained on the scaled data relative to that of the standard data. \textbf{Ratio:} The ratio of the change value to the accuracy of the standard data. \textbf{Average (abs)} and \textbf{Standard deviation (abs)} are calculated from the absolute values of change and ratio of 22 models.)}\label{cross_val_compare}
\centering
\scalebox{0.75}{
\begin{tabular}{@{} cccc@{} }
\toprule
 Models            & Up/down & Change (\%) & Ratio (\%) \\
\midrule
VGG11                   & down & $-$1.90 & $-$2.25 \\
VGG13                   & down & $-$1.61 & $-$1.86 \\
VGG16                   & down & $-$3.13 & $-$3.80 \\
VGG19                   & down & $-$4.61 & $-$5.43 \\
ResNet18                & down & $-$1.09 & $-$1.24 \\
ResNet34                & down & $-$0.56 & $-$0.65 \\
ResNet50                & down & \textbf{$-$5.00} & \textbf{$-$5.73} \\
ResNet101               & down & $-$4.41 & $-$5.11 \\
DenseNet121           & down & $-$3.23 & $-$3.73 \\
DenseNet169           & down & $-$1.01 & $-$1.17 \\
InceptionV3             & down & $-$2.50 & $-$2.90\\
Xception                  & down & $-$0.88 & $-$0.99 \\
AlexNet                    & down & $-$4.03 & $-$4.73 \\
GoogLeNet               & down & $-$2.32 & $-$2.65 \\
MobileNetV2             & down & $-$0.39 & $-$0.46 \\
ShuffleNetV2$\times$1.0        & up   & $+$0.03  & $+$0.03 \\
ShuffleNetV2$\times$0.5        & down & $-$0.57 & $-$0.68 \\
InceptionResNetV1       & down & $-$1.93 & $-$2.21 \\
ViT                          & up   & \textbf{$+$2.07}  & \textbf{$+$2.55}  \\
BoTNet                    & down & $-$3.43 & $-$4.21 \\
DeiT                        & up   & $+$0.07 & $+$0.08  \\
T2T-ViT                   & down & $-$1.55 & $-$1.82 \\
Average                  & down & $-$1.91 & $-$2.23 \\
Average (abs)           & -    & 2.11  & 2.47  \\
Standard deviation (abs) & -    & 1.48  & 1.74	\\
\bottomrule
\end{tabular}}
\end{table}

The classification performance of the models of the five-fold cross-validation experiment on the SIPaKMeD dataset is presented in Table~\ref{cross_val_val}. In this table, among all CNNs trained on scaled data, Xception exhibits the highest classification accuracy of 87.95\%. Among all VTs trained on scaled data, DeiT exhibits the highest accuracy of 85.16\%. With CNNs trained on standard data, the accuracy of Xception is still the highest at 88.83\%. Among all VTs trained on standard data, T2T-ViT exhibits the highest accuracy.

A comparison of the validation results between the scaled data and standard data of the five-fold cross-validation experiment on the SIPaKMeD dataset is shown in Table~\ref{cross_val_compare}. In this table, the accuracies of three models, including ShuffleNetV2$\times$1.0, ViT, and DeiT, trained on scaled data are higher than that of standard data. The accuracies of 19 models, including VGG11, VGG13, VGG16, VGG19, ResNet18, ResNet34, ResNet50, ResNet101, DenseNet121, DenseNet169, InceptionV3, Xception, AlexNet, GoogLeNet, MobileNetV2, ShuffleNetV2$\times$0.5, InceptionResNetV1, BoTNeT, and T2T-ViT, trained on scaled data are lower than that of the standard data. The model with the highest increase in accuracy is ViT (2.07\%) and the rate of increase is 2.55\%. The decrease in the accuracy of ResNet50 is the highest and corresponds to 5.00\% and the ratio is 5.73\%. After the calculation, the average change in accuracy is -1.91\%. The average of the absolute value of the change in accuracy is 2.11\%, and the standard deviation of the absolute value of changes in accuracy is 1.48\%. The average ratio of changes in accuracy is -2.23\%. The average of the absolute value of the change ratios is 2.47\%, and the standard deviation is 1.74\%.

Hence, overall the accuracies of some models increase and some decrease. Moreover, the fluctuations are not very high. Table~\ref{cross_val_compare} shows that after using the five-fold cross-validation method, when compared with Table~\ref{tbl6}, the performance of ResNet, DenseNet, Xception, and GoogLeNet on the scaled data are lower than that on the standard data. The performance of the remaining models remains the same. Hence, this indicates that the deep learning method is robust when the aspect ratios of cells change.

\subsubsection{Extended experiment on SIPaKMeD dataset: classification with data augmentation\label{sec:SIPaKMeD_aug_extend}}

The classification performance on the test set of SIPaKMeD-augmented data is shown in Table~\ref{tbl9}. In this table, in CNNs trained on scaled data, ResNet18 exhibits the highest accuracy, which is 97.15\%. In VTs trained on scaled data, ViT exhibits the highest accuracy rate of 95.54\%. In CNNs trained on standard data, GoogLeNet exhibits the highest accuracy of 97.40\%. In VTs trained on standard data, DeiT exhibits the highest accuracy rate.

\begin{table*}
\caption{Classification performance of models on the \textbf{test set} of SIPaKMeD augmented data.}\label{tbl9}
\centering
\scalebox{0.72}{
\begin{tabular}{@{} ccccccccccc@{} }
\toprule
 &\multicolumn{4}{c}{Scaled data} &  & & \multicolumn{4}{c}{Standard data} \\
\midrule
Models           & Avg. P (\%) & Avg. R (\%)& Avg. F1 (\%) & Accuracy (\%) &&& Avg. P (\%) & Avg. R (\%) &Avg. F1 (\%) &  Accuracy (\%)\\
\midrule
VGG11             & 96.20 & 96.20 & 96.10 & 96.03 &  &  & 95.20 & 94.80 & 94.90 & 94.80 \\
VGG13             & 94.70 & 94.60 & 94.60 & 94.55 &  &  & 96.30 & 96.30 & 96.30 & 96.28 \\
VGG16             & 94.10 & 93.90 & 94.00 & 93.93 &  &  & 95.30 & 95.10 & 95.20 & 95.17 \\
VGG19             & 93.10 & 92.50 & 92.70 & 92.57 &  &  & 95.30 & 95.20 & 95.20 & 95.17 \\
ResNet18          & 97.20 & 97.20 & 97.20 & \textbf{97.15} &  &  & 96.90 & 96.80 & 96.80 & 96.78 \\
ResNet34          & 96.30 & 96.20 & 96.20 & 96.16 &  &  & 96.10 & 96.00 & 96.10 & 96.03 \\
ResNet50          & 95.60 & 95.60 & 95.60 & 95.54 &  &  & 96.20 & 96.20 & 96.10 & 96.03 \\
ResNet101         & 96.00 & 95.90 & 95.90 & 95.91 &  &  & 95.80 & 95.70 & 95.70 & 95.66 \\
DenseNet121       & 96.90 & 96.90 & 96.90 & 96.90 &  &  & 97.10 & 97.10 & 97.00 & 97.02 \\
DenseNet169       & 96.30 & 96.20 & 92.00 & 96.16 &  &  & 96.60 & 96.50 & 96.50 & 96.53 \\
InceptionV3       & 95.40 & 95.20 & 95.20 & 95.17 &  &  & 96.80 & 96.80 & 96.80 & 96.78 \\
Xception          & 97.00 & 97.10 & 97.00 & 97.02 &  &  & 96.10 & 96.10 & 95.80 & 95.79 \\
AlexNet           & 95.10 & 94.90 & 94.90 & 94.92 &  &  & 95.60 & 95.70 & 95.70 & 95.66 \\
GoogLeNet         & 96.80 & 96.80 & 96.80 & 96.78 &  &  & 97.50 & 97.40 & 97.40 & \textbf{97.40} \\
MobileNetV2       & 96.50 & 96.60 & 96.50 & 96.53 &  &  & 96.80 & 96.80 & 96.80 & 96.78 \\
ShuffleNetV2$\times$1.0  & 96.90 & 96.80 & 96.80 & 96.78 &  &  & 96.10 & 96.10 & 96.10 & 96.03 \\
ShuffleNetV2$\times$0.5  & 95.30 & 95.20 & 95.20 & 95.17 &  &  & 95.90 & 95.80 & 95.80 & 95.79 \\
InceptionResNetV1 & 95.10 & 95.10 & 95.00 & 95.04 &  &  & 97.20 & 97.20 & 97.20 & 97.15 \\
ViT               & 95.80 & 95.60 & 95.60 & \textbf{95.54} &  &  & 93.10 & 93.00 & 93.00 & 92.94 \\
BoTNet            & 87.60 & 87.60 & 87.50 & 87.62 &  &  & 88.90 & 89.00 & 88.90 & 88.98 \\
DeiT              & 95.10 & 95.00 & 95.00 & 94.92 &  &  & 95.00 & 94.90 & 95.00 & \textbf{94.92} \\
T2T-ViT           & 94.50 & 94.50 & 94.40 & 94.43 &  &  & 92.20 & 92.10 & 92.10 & 92.07\\
\bottomrule
\end{tabular}}
\end{table*}

\begin{table}
\caption{Comparison table of test results between the augmented scaled data and standard data on SIPaKMeD dataset. (\textbf{Up/down:} The accuracy of the model trained on the scaled data is higher/lower than that of the standard data. \textbf{Change:} The change value of the accuracy of the model trained on the scaled data relative to that of the standard data. \textbf{Ratio:} The ratio of the change value to the accuracy of the standard data. \textbf{Average (abs)} and \textbf{Standard deviation (abs)} are calculated from the absolute values of change and ratio of 22 models.)}\label{tbl10}
\centering
\scalebox{0.75}{
\begin{tabular}{@{} cccc@{} }
\toprule
 Models            & Up/down & Change (\%) & Ratio (\%) \\
\midrule
VGG11             & up    & $+$1.23  & $+$1.30  \\
VGG13             & down  & $-$1.73 & $-$1.80 \\
VGG16             & down  & $-$1.24 & $-$1.30 \\
VGG19             & down  & \textbf{$-$2.60} & \textbf{$-$2.73} \\
ResNet18          & up    & $+$0.37  & $+$0.38  \\
ResNet34         & up    & $+$0.13  & $+$0.14  \\
ResNet50          & down  & $-$0.49 & $-$0.51 \\
ResNet101         & up    & $+$0.25  & $+$0.26  \\
DenseNet121       & down  & $-$0.12 & $-$0.12 \\
DenseNet169      & down  & $-$0.37 & $-$0.38 \\
InceptionV3       & down  & $-$1.61 & $-$1.66 \\
Xception          & up    & $+$1.23  & $+$1.28  \\
AlexNet           & down  & $-$0.74 & $-$0.77 \\
GoogLeNet         & down  & $-$0.62 & $-$0.64 \\
MobileNetV2       & down  & $-$0.25 & $-$0.26 \\
ShuffleNetV2$\times$1.0  & up    & $+$0.75  & $+$0.78  \\
ShuffleNetV2$\times$0.5  & down  & $-$0.62 & $-$0.65 \\
InceptionResNetV1  & down  & $-$2.11 & $-$2.17 \\
ViT               & up    & \textbf{$+$2.60}  & \textbf{$+$2.80}  \\
BoTNet         & down  & $-$1.36 & $-$1.53 \\
DeiT             & equal & 0.00  & 0.00  \\
T2T-ViT        & up    & $+$2.36  & $+$2.56  \\ 
Average          &  down   & $-$0.22  & $-$0.23  \\
Average (abs)    &   -  & 1.04  & 1.09  \\ 	 		 	
Standard deviation (abs) & -  & 0.81   &  	0.86   \\ 	 	 	 	 	 
\bottomrule
\end{tabular}}
\end{table}

Table~\ref{tbl10} compares the accuracy of the test set using scaled data and standard data. The accuracies of eight models, including VGG11, ResNet18, ResNet34, ResNet101, Xception, ShuffleNetV2$\times$1.0, ViT, and T2T-ViT, trained on scaled data are higher than that of standard data. The accuracies of 13 models, including VGG13, VGG16, VGG19, ResNet50, DenseNet121, DenseNet169, InceptionV3, AlexNet, GoogLeNet, MobileNetV2, ShuffleNetV2$\times$0.5, InceptionResNetV1, and BoTNet are reduced, and the accuracy of DeiT is equal. The model with the highest improvement is ViT, which exhibits an increase of 2.60\% and an increase rate of 2.80\%. The decrease in the accuracy of VGG19 is the highest and corresponds to 2.60\%, and the ratio is 2.73\%. After the calculation, the average change in accuracy is -0.22\%. The average absolute value of the change in accuracy is 1.04\%, and the standard deviation of the absolute value of the changes is 0.81\%. The average ratio of changes in accuracy is -0.23\%. The average of the absolute value of the change ratios is 1.09\%, and the standard deviation is 0.86\%.

Overall, the accuracies of certain models increase and that of certain models decrease. Table~\ref{tbl10} shows that after data augmentation, when compared with Table~\ref{tbl6}, the performance of the DenseNet and Lightweight networks on the scaled data is lower than that of the standard data. The remaining models exhibit the same performance. Therefore, after augmenting the training data, some redundant information appears, resulting in a negative impact on certain models. However, the results indicate that the deep learning method is highly robust to changes in the aspect ratio of cells in cervical cytopathological images.

\subsubsection{Extended experiment on Herlev dataset\label{Herlev_extend}}

\begin{table*}
\caption{Two-class classification performance of models on the \textbf{test set} of Herlev augmented data.}\label{tbl13}
\centering
\scalebox{0.72}{
\begin{tabular}{@{} ccccccccccc@{} }
\toprule
 &\multicolumn{4}{c}{Scaled data} &  & & \multicolumn{4}{c}{Standard data} \\
\midrule
Models           & Avg. P (\%) & Avg. R (\%)& Avg. F1 (\%) & Accuracy (\%) &&& Avg. P (\%) & Avg. R (\%) &Avg. F1 (\%) &  Accuracy (\%)\\
\midrule
VGG11             & 86.00 & 81.80 & 83.60 & 87.97 &  &  & 85.50 & 80.80 & 82.70 & 87.43 \\
VGG13             & 89.90 & 81.20 & 84.30 & 89.07 &  &  & 90.70 & 87.70 & 89.00 & 91.80 \\
VGG16             & 86.10 & 80.10 & 82.40 & 87.43 &  &  & 86.60 & 84.50 & 85.50 & 89.07 \\
VGG19             & 36.90 & 50.00 & 42.40 & 73.77 &  &  & 90.50 & 79.40 & 83.00 & 88.52 \\
ResNet18          & 93.20 & 86.80 & 89.40 & 92.34 &  &  & 91.50 & 89.80 & 90.60 & 92.89 \\
ResNet34          & 92.40 & 84.70 & 87.60 & 91.25 &  &  & 88.10 & 86.00 & 86.90 & 90.16 \\
ResNet50          & 89.10 & 88.00 & 88.60 & 91.25 &  &  & 93.20 & 86.80 & 89.40 & 92.34 \\
ResNet101         & 91.00 & 88.70 & 89.80 & 92.34 &  &  & 88.70 & 91.10 & 89.80 & 91.80 \\
DenseNet121       & 91.70 & 88.10 & 89.70 & 92.34 &  &  & 96.00 & 91.30 & 93.40 & \textbf{95.08} \\
DenseNet169       & 91.50 & 89.80 & 90.60 & 92.89 &  &  & 89.60 & 89.10 & 89.30 & 91.80 \\
InceptionV3       & 87.50 & 84.90 & 86.10 & 89.61 &  &  & 89.70 & 85.60 & 87.40 & 90.71 \\
Xception          & 92.80 & 88.50 & 90.40 & 92.89 &  &  & 89.10 & 88.00 & 88.60 & 91.25 \\
AlexNet           & 91.60 & 85.30 & 87.80 & 91.25 &  &  & 86.50 & 82.80 & 84.40 & 88.52 \\
GoogLeNet         & 91.10 & 92.20 & 91.60 & \textbf{93.44} &  &  & 90.70 & 83.30 & 86.10 & 90.16 \\
MobileNetV2       & 90.40 & 85.00 & 87.10 & 90.71 &  &  & 89.00 & 81.90 & 84.60 & 89.07 \\
ShuffleNetV2$\times$1.0  & 92.40 & 84.70 & 87.60 & 91.25 &  &  & 90.70 & 83.30 & 86.10 & 90.16 \\
ShuffleNetV2$\times$0.5  & 90.60 & 89.40 & 90.00 & 92.34 &  &  & 90.50 & 79.40 & 83.00 & 88.52 \\
InceptionResNetV1 & 90.80 & 86.00 & 88.00 & 91.25 &  &  & 88.60 & 87.00 & 87.80 & 90.71 \\
ViT               & 89.70 & 90.80 & 90.20 & 92.34 &  &  & 86.80 & 73.90 & 77.40 & 85.24 \\
BoTNet            & 91.30 & 81.60 & 84.90 & 89.61 &  &  & 90.10 & 78.40 & 82.00 & 87.97 \\
DeiT              & 91.50 & 91.50 & 91.60 & \textbf{93.44} &  &  & 88.50 & 78.00 & 81.40 & 87.43 \\
T2T-ViT           & 84.70 & 83.80 & 84.20 & 87.90 &  &  & 91.70 & 82.60 & 85.90 & \textbf{90.16} \\
\bottomrule
\end{tabular}}
\end{table*}

\begin{table}
\caption{Comparison table of test results between the augmented scaled data and standard data on Herlev dataset. (\textbf{Up/down:} The accuracy of the model trained on the scaled data is higher/lower than that of the standard data. \textbf{Change:} The change value of the accuracy of the model trained on the scaled data relative to that of the standard data. \textbf{Ratio:} The ratio of the change value to the accuracy of the standard data. \textbf{Average (abs)} and \textbf{Standard deviation (abs)} are calculated from the absolute values of change and ratio of 22 models.)}\label{tbl14}
\centering
\scalebox{0.75}{
\begin{tabular}{@{} cccc@{} }
\toprule
 Models            & Up/down & Change (\%) & Ratio (\%) \\
\midrule
VGG11             & up   & $+$0.54   & $+$0.62   \\
VGG13             & down & $-$2.73  & $-$2.97  \\
VGG16             & down & $-$1.64  & $-$1.84 \\
VGG19             & down & \textbf{$-$14.75\%} & \textbf{$-$16.66} \\
ResNet18          & down & $-$0.55  & $-$0.59  \\
ResNet34          & up   & $+$1.09   & $+$1.21   \\
ResNet50          & down & $-$1.09  & $-$1.18  \\
ResNet101        & up   & $+$0.54   & $+$0.59   \\
DenseNet121    & down & $-$2.74  & $-$2.88  \\
DenseNet169    & up   & $+$1.09  & $+$1.19   \\
InceptionV3      & down & $-$1.10  & $-$1.21  \\
Xception           & up   & $+$1.64   & $+$1.80   \\
AlexNet             & up   & $+$2.73   & $+$3.08   \\
GoogLeNet       & up   & $+$3.28  & $+$3.64   \\
MobileNetV2      & up   & $+$1.64   & $+$1.84  \\
ShuffleNetV2$\times$1.0  & up   & $+$1.09   & $+$1.21   \\
ShuffleNetV2$\times$0.5  & up   & $+$3.82   & $+$4.32   \\
InceptionResNetV1 & up   & $+$0.54   & $+$0.60   \\
ViT               & up   & \textbf{$+$7.10}   & \textbf{$+$8.33}   \\
BoTNet         & up   & $+$1.64   & $+$1.86   \\
DeiT             & up   & $+$6.01   & $+$6.87   \\
T2T-ViT         & down & $-$2.26  & $-$2.51 \\
Average            &  up   & $+$0.27  & $+$0.33  \\
Average (abs)    &   -  & 2.71  & 3.05   \\ 	 		 	
Standard deviation (abs) & -  & 3.11   &  	3.55   \\ 	
\bottomrule
\end{tabular}}
\end{table}

The two-class classification performance on the test set of Herlev-augmented data is presented in Table~\ref{tbl13}. In this table, in CNNs trained on scaled data, GoogLeNet exhibits the highest classification accuracy of 93.44\%. Among all VTs trained on scaled data, DeiT exhibits the highest accuracy of 93.44\%. Among the CNNs trained on standard data, DenseNet121 exhibits the highest accuracy of 95.08\%. In VTs trained on standard data, T2T-ViT exhibits the highest accuracy.

Table~\ref{tbl14} compares the accuracy of the test set using scaled data and standard data. The accuracies of 14 models, including VGG11, ResNet34, ResNet101, DenseNet169, Xception, AlexNet, GoogLeNet, MobileNetV2, ShuffleNetV2$\times$1.0, ShuffleNetV2$\times$0.5, InceptionResNetV1, ViT, BoTNet, and DeiT, trained on scaled data are higher than those of the standard data, and the accuracies of eight models, including VGG13, VGG16, VGG19, ResNet18, ResNet50, DenseNet121, InceptionV3, and T2T-ViT are reduced. The model with the highest improvement is ViT, and it exhibits an increase of 7.10\% and a ratio of 8.33\%. The accuracy of VGG19 is reduced by 14.75\% with a ratio of 16.66\%. After the calculation, the average change in accuracy is +0.27\%. The average absolute value of the change in accuracy is 2.71\%, and the standard deviation of the absolute value of the changes is 3.11\%. The average ratio of changes in accuracy is +0.33\%. The average absolute value of the change ratio is 3.05\%, and the standard deviation is 3.55\%.

Overall, the trends indicate that the accuracies of certain models increase and that of certain models decrease. Given that a small amount of data is used in the Herlev dataset and high class imbalance of the two categories, the classification performance is reduced when compared with the SIPaKMeD experiment. However, overall, the accuracies of the models do not fluctuate significantly. On the Herlev dataset, changes in the aspect ratio of cells in cervical cell images do not affect the robustness of deep learning methods.

\subsubsection{Extended experiment on Cropped data\label{random_crop}}

\begin{figure}[htbp]
	\centering
	  \includegraphics[scale=0.7]{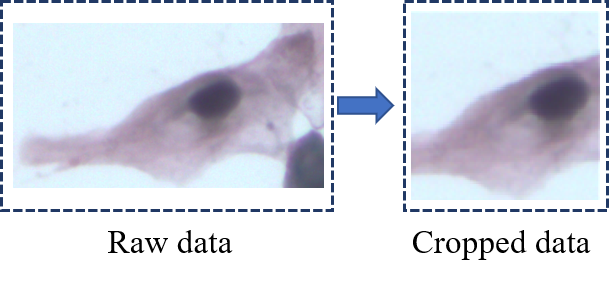}
	\caption{Example of cropping pre-process.}
	\label{FIG:crop_data}
\end{figure}

To demonstrate this conclusion, we use another preprocessing method, which is similar to standardisation, to pre-process the image as shown in Fig.~\ref{FIG:crop_data}. First, the raw data dimension is proportionally resized such that the smallest side corresponds to 224 pixels. Then a random crop is applied to the resized data to generate  images of 224 $\times$ 224 pixels. The pre-processed data are termed as ``cropped data". The cropped data are input into models for training and testing, and the data is compared with the test results of the scaled data.

The performance of the models on the scaled data and cropped data test sets of the SIPaKMeD dataset is shown in Table~\ref{cropped_data_performance}. In this table, GoogLeNet exhibits the highest accuracy of 96.03\% in CNNs trained on scaled data. In VTs trained on scaled data, DeiT exhibits the highest accuracy rate of 95.42\%. In CNNs trained on cropped data, Xception exhibits the highest accuracy of 95.42\%. In VTs trained on cropped data, DeiT exhibits the highest accuracy rate.

Table~\ref{cropped_data_compare} compares the accuracy of the test set using scaled data and cropped data. The accuracies of 17 models, including VGG11, VGG13, VGG16, VGG19, ResNet18, ResNet34, DenseNet121, DenseNet169, Xception, GoogLeNet, MobileNetV2, ShuffleNetV2$\times$1.0, ShuffleNetV2$\times$0.5, InceptionResNetV1, ViT, DeiT, and T2T-ViT, trained on scaled data are higher than that of the cropped data, and the accuracies of the five models, ResNet50, ResNet101, InceptionV3, AlexNet, and BoTNet are reduced. The model with the highest improvement is DenseNet169, and it exhibits an increase of 2.84\% and a ratio of 3.06\%. The decrease in the accuracy of BoTNet is the highest (4.46\%) with a ratio of 5.00\%. After the calculation, the average change in accuracy is +0.65\%. The average absolute value of the change in accuracy is 1.14\%, and the standard deviation of the absolute value of changes is 1.04\%. The average ratio of changes in accuracy is +0.71\%. The average absolute value of the change ratios is 1.26\%, and the standard deviation is 1.17\%.

Overall trends indicate that the accuracies of certain models increase and that of certain models decrease. We use cropped data as opposed to standard data to conduct experiments, and the results indicate that the deep learning method is very robust to changes in the aspect ratio of cells in cervical cytopathological images.

\begin{table*}
\caption{Classification performance of models on the \textbf{scaled data and cropped data test sets} of SIPaKMeD dataset.}\label{cropped_data_performance}
\centering
\scalebox{0.72}{
\begin{tabular}{@{} ccccccccccc@{} }
\toprule
 &\multicolumn{4}{c}{Scaled data} &  & & \multicolumn{4}{c}{Cropped data} \\
\midrule
Models           & Avg. P (\%) & Avg. R (\%)& Avg. F1 (\%) & Accuracy (\%) &&& Avg. P (\%) & Avg. R (\%) &Avg. F1 (\%) &  Accuracy (\%)\\
\midrule
VGG11             & 93.70 & 93.60 & 93.60 & 93.56 &  &  & 92.80 & 92.60 & 92.60 & 92.57 \\
VGG13             & 91.60 & 91.20 & 91.30 & 91.21 &  &  & 90.90 & 90.20 & 90.40 & 90.22 \\
VGG16             & 89.80 & 89.20 & 89.30 & 89.23 &  &  & 88.70 & 87.30 & 87.50 & 87.38 \\
VGG19             & 88.60 & 88.10 & 88.30 & 88.11 &  &  & 86.20 & 85.90 & 85.90 & 85.89 \\
ResNet18          & 95.60 & 95.50 & 95.60 & 95.54 &  &  & 95.20 & 95.20 & 95.20 & 95.17 \\
ResNet34          & 94.30 & 94.00 & 94.10 & 94.05 &  &  & 94.20 & 93.90 & 94.00 & 93.94 \\
ResNet50          & 93.10 & 92.80 & 92.80 & 92.82 &  &  & 93.60 & 93.50 & 93.50 & 93.44 \\
ResNet101         & 92.30 & 92.00 & 92.00 & 91.95 &  &  & 92.20 & 92.10 & 92.10 & 92.08 \\
DenseNet121       & 95.80 & 95.70 & 95.70 & 95.66 &  &  & 94.20 & 94.10 & 94.10 & 94.06 \\
DenseNet169       & 95.80 & 95.70 & 95.70 & 95.66 &  &  & 93.20 & 92.80 & 92.90 & 92.82 \\
InceptionV3       & 93.20 & 93.00 & 93.10 & 92.94 &  &  & 93.30 & 93.10 & 93.10 & 93.07 \\
Xception          & 96.10 & 95.90 & 95.90 & 95.91 &  &  & 95.50 & 95.40 & 95.40 & \textbf{95.42} \\
AlexNet           & 92.00 & 91.90 & 91.90 & 91.83 &  &  & 92.30 & 92.00 & 92.10 & 91.96 \\
GoogLeNet         & 96.30 & 96.00 & 96.10 & \textbf{96.03} &  &  & 95.00 & 94.90 & 94.90 & 94.93 \\
MobileNetV2       & 93.80 & 93.70 & 93.70 & 93.68 &  &  & 93.10 & 92.40 & 92.60 & 92.45 \\
ShuffleNetV2$\times$1.0  & 92.30 & 91.90 & 91.90 & 91.83 &  &  & 91.40 & 91.00 & 91.10 & 90.97 \\
ShuffleNetV2$\times$0.5  & 92.70 & 92.30 & 92.40 & 92.32 &  &  & 90.80 & 90.50 & 90.50 & 90.47 \\
InceptionResNetV1 & 93.80 & 93.70 & 93.70 & 93.68 &  &  & 93.10 & 92.80 & 92.90 & 92.82 \\
ViT               & 93.60 & 93.60 & 93.60 & 93.56 &  &  & 92.10 & 91.70 & 91.80 & 91.71 \\
BoTNet            & 85.80 & 84.80 & 84.90 & 84.77 &  &  & 89.60 & 89.30 & 89.30 & 89.23 \\
DeiT              & 95.50 & 95.40 & 95.50 & \textbf{95.42} &  &  & 95.20 & 95.20 & 95.20 & \textbf{95.17} \\
T2T-ViT           & 91.90 & 91.70 & 91.70 & 91.70 &  &  & 92.20 & 91.40 & 91.60 & 91.46\\
\bottomrule
\end{tabular}}
\end{table*}

\begin{table}
\caption{Comparison table of test results between the scaled data and cropped data on SIPaKMeD dataset. (\textbf{Up/down:} The accuracy of the model trained on the scaled data is higher/lower than that of the cropped data. \textbf{Change:} The change value of the accuracy of the model trained on the scaled data relative to that of the cropped data. \textbf{Ratio:} The ratio of the change value to the accuracy of the cropped data. \textbf{Average (abs)} and \textbf{Standard deviation (abs)} are calculated from the absolute values of change and ratio of 22 models.)}\label{cropped_data_compare}
\centering
\scalebox{0.75}{
\begin{tabular}{@{} cccc@{} }
\toprule
 Models            & Up/down & Change (\%) & Ratio (\%) \\
\midrule
VGG11                    & up   & $+$0.99  & $+$1.07  \\
VGG13                    & up   & $+$0.99  & $+$1.10  \\
VGG16                    & up   & $+$1.85  & $+$2.12  \\
VGG19                    & up   & $+$2.22  & $+$2.58  \\
ResNet18                 & up   & $+$0.37  & $+$0.39  \\
ResNet34                 & up   & $+$0.11  & $+$0.12  \\
ResNet50                 & down & $-$0.62 &$-$0.66 \\
ResNet101                & down & $-$0.13 & $-$0.14 \\
DenseNet121              & up   & $+$1.60  & $+$1.70  \\
DenseNet169              & up   & \textbf{$+$2.84}  & \textbf{$+$3.06}  \\
InceptionV3              & down & $-$0.13 & $-$0.14 \\
Xception                 & up   & $+$0.49  & $+$0.51  \\
AlexNet                  & down & $-$0.13 & $-$0.14 \\
GoogLeNet                & up   & $+$1.10  & $+$1.16  \\
MobileNetV2              & up   & $+$1.23  & $+$1.33  \\
ShuffleNetV2$\times$1.0         & up   & $+$0.86  & $+$0.95  \\
ShuffleNetV2$\times$0.5         & up   & $+$1.85  & $+$2.04  \\
InceptionResNetV1        & up   & $+$0.86  & $+$0.93  \\
ViT                      & up   & $+$1.85  & $+$2.02  \\
BoTNet                   & down & \textbf{$-$4.46} & \textbf{$-$5.00} \\
DeiT                     & up   & $+$0.25  & $+$0.26  \\
T2T-ViT                  & up   & $+$0.24  & $+$0.26  \\
Average                  & up   & $+$0.65  & $+$0.71  \\
Average (abs)            & -    & 1.14  & 1.26  \\
Standard deviation (abs) & -    & 1.04  & 1.17\\
\bottomrule
\end{tabular}}
\end{table}

\section{Conclusion and future work\label{sec:Conclusion and Future Work}}

The image input required by the deep learning model is generally consistent, but the dimensions of many clinical medical images are inconsistent. The aspect ratios of the images are affected when they are directly resized. Clinically, the aspect ratios of cells inside cytopathological images provide important information to doctors for diagnosing cancer. Therefore, it is difficult to directly resize them. However, many existing studies have directly resized images, and the results are still robust. To determine a reasonable explanation, cervical cells are used as an example to test the robustness of the deep learning method. First, the raw data are standardised and resized directly to obtain the standard and scaled data, respectively. Then, the standard and scaled data are resized to 224 $\times$ 224 pixels. Finally, 22 deep learning models are used to classify the standard data and scaled data. The experimental results indicate that in the case of classifying the scaled data without aspect ratio information, the classification effects of some models are slightly lower than that of the standard data. The decrease in the classification effects is not high. However, the effects of some models are improved. Overall, the results indicate that the deep learning method is robust to changes in the aspect ratio of cells in cervical cytopathological images. This is because although aspect ratios of cells change, the main visual information that distinguishes the cells is concentrated on the nucleus. The details of the nuclei of different cells differ, and the signals appear as high-frequency signals with their own characteristics. This provides good features for image classification. The limitation of this study is that the input images must be cytopathological images that contain cell aspect ratios.

Recently, the popular Transformer surpasses the CNN network in the field of computer vision. In this study, the Transformer does not exhibit superior results, and BoTNet combined with the Transformer and ResNet does not perform well on cervical cell images. We assume that combining Xception, which performs well in this study, with Transformer can lead to better performance. The Transformer focuses on global information. Hence, we can fuse the Transformer and classical local features to obtain better classification performance. In the future, we can try to use new VTs to classify medical images and make improvements and supplements based on the structure of the model. Additionally, standardisation and centralisation can be used for data preprocessing, which can increase the accuracy of the model.

\section*{Acknowledgements}
This work is supported by the ``National Natural Science Foundation of China'' (No.61806047) 
and the ``Fundamental Research Funds for the Central Universities'' (No. N2019003). 
We thank Miss Zixian Li and Mr. Guoxian Li for their important discussion. 

\section*{Declaration of competing interest} 
The authors declare that they have no conflict of interest.

\bibliographystyle{elsarticle-num}
\bibliography{Wanli}

\onecolumn

\begin{appendices}
\section*{Appendix}

\begin{table*}[h]
\centering
\caption{A brief introduction to the deep learning models used in the robustness comparison experiment.}\label{model_introduce}
\scalebox{0.9}{
\begin{tabular}{@{} ccccl@{} }
\toprule
Model & Ref. & Parameters & Top-1 accuracy & Major remarks \\
\midrule
VGG11 &  \cite{simonyan2014very} & 133M & 69.02\% & a.The network has deeper layers and uses \\
VGG13 &   & 133M & 69.93\% & \quad  smaller filters. \\
VGG16 &   & 138M & 71.59\%  & b.Training takes a long time. \\
VGG19 &   & 144M & 72.38\% &  \\

ResNet18 & \cite{he2016deep} & 11.7M & 69.76\% & a.The residual block is introduced. \\
ResNet34 &   & 21.8M & 73.31\% & b.The problem of rapid decline in accuracy\\
ResNet50 &   & 25.6M & 76.13\%  &   \quad due to the increase in the number of\\
ResNet101 &   &  44.7M & 77.37\% &  \quad network layers is solved. \\

DenseNet121 & \cite{huang2017densely} &  8M & 74.43\% & a.A direct connection between any two  \\
DenseNet169 &  & 14.3M & 75.60\% &  \quad layers with the same feature map size  \\
&  &  &  &  \quad is introduced. \\
&  &  &  &  b.It can scale to hundreds of layers. \\

InceptionV3 & \cite{Szegedy2016Rethinking} &  23.8M & 77.29\% & a.It uses improved Inception Module. \\
&  &  &  &  b.It introduces the idea of Factorization  \\
&  &  &  &  \quad into small convolutions. \\

Xception & \cite{Chollet2017Xception} & 22.9M & 79.00\% & a.It is an improved version of InceptionV3. \\
&  &  &  &  b.It uses depthwise separable convolutions.  \\

AlexNet & \cite{russakovsky2015imagenet} & 60M & 56.52\% &  a.ReLu activation function is introduced. \\
 &  &  &  &  b.The local response normalization   \\
 &  &  &  & \quad layer is proposed to improve the  \\
 &  &  &  & \quad generalisation ability of the model. \\

GoogLeNet & \cite{Szegedy2015googlenet} & 6.8M & 69.78\% & a.It introduces the inception structure. \\
&  & &  & b.It adopts multi-scale processing and\\
&  & &  & \quad Hebbian principle. \\

MobileNetV2 & \cite{sandler2018mobilenetv2} & 3.4M & 71.89\% & a.Inverted residual with linear bottleneck \\
&  & &  & \quad  is introduced. \\

ShuffleNetV2$\times$1.0 & \cite{ma2018shufflenet} & 2.3M & 69.36\% & a.It uses channel split at the beginning of  \\
ShuffleNetV2$\times$0.5 &  & 1.4M & 60.55\% & \quad the block.\\

InceptionResNetV1 & \cite{szegedy2017inception} & 8M & 78.70\% & a.It combines the Inception block and the  \\
&  & &  & \quad ResNet connection. \\

ViT & \cite{Dosovitskiy2020An} & 86M & 73.38\% & a.It applies the Transformer to the field of  \\
&  & &  & \quad computer vision. \\
&  &  &  &  b.It is better than CNN under the training  \\
&  & &  & \quad of a large amount of data. \\

BoTNet & \cite{Srinivas2021Bottleneck} & 20.8M & 77.00\% & a.It combines self-attention with ResNet.  \\

DeiT & \cite{touvron2020training} & 5M & 72.20\% & a.Distillation token is introduced.\\
&  &  &  &  b.The training of DeiT requires less data   \\
&  & &  & \quad and computing resources. \\

T2T-ViT & \cite{yuan2021tokens} &  4.2M & 71.70\% & a.A new T2T mechanism based on ViT \\
&  & &  & \quad is used. \\
\bottomrule
\end{tabular}}
\end{table*}

\end{appendices}

\end{document}